\documentclass[10pt,journal,compsoc]{IEEEtran}
\usepackage{hyperref}
\usepackage{graphicx}
\usepackage[flushleft]{threeparttable}
\usepackage{booktabs} 
\usepackage[caption=false]{subfig}
\usepackage[center]{caption}
\usepackage{amsfonts}
\usepackage{multirow}
\usepackage{tabularx}
\usepackage{xcolor}
\usepackage{amsmath}

\usepackage{algorithm}
\usepackage{algpseudocode}

\newcolumntype{M}{>{$}c<{$}}
\newcolumntype{C}{>{\centering\arraybackslash}p}
\newcolumntype{Y}{>{\centering\arraybackslash}X}

\newcommand{\specialcell}[2][c]{%
  \begin{tabular}[#1]{@{}c@{}}#2\end{tabular}}

\newcommand{\leftcell}[2][l]{%
  \begin{tabular}[#1]{@{}l@{}}#2\end{tabular}}

\ifCLASSOPTIONcompsoc
  \usepackage[nocompress]{cite}
\else
  \usepackage{cite}
\fi

\begin{document}
\title{AFR-Net: Attention-Driven Fingerprint Recognition Network}
%
%
%
%

\author{Steven~A.~Grosz and~Anil~K.~Jain,~\IEEEmembership{Life~Fellow,~IEEE}
\IEEEcompsocitemizethanks{\IEEEcompsocthanksitem S.A. Grosz and A.K. Jain are with the Department of Computer Science and Engineering, Michigan State University, East Lansing, MI, 48824 USA (e-mail: groszste@cse.msu.edu, jain@cse.msu.edu).}
}

\markboth{Journal of \LaTeX\ Class Files,~Vol.~14, No.~8, August~2015}%
{Grosz and Jain: AFR-Net: Attention-Driven Fingerprint Recognition Network}
%



\IEEEtitleabstractindextext{%
\begin{abstract}
The use of vision transformers (ViT) in computer vision is increasing due to limited inductive biases (e.g., locality, weight sharing, etc.) and increased scalability compared to other deep learning methods. This has led to some initial studies on the use of ViT for biometric recognition, including fingerprint recognition. In this work, we improve on these initial studies for transformers in fingerprint recognition by i.) evaluating additional attention-based architectures, ii.) scaling to larger and more diverse training and evaluation datasets, and iii.) combining the complimentary representations of attention-based and CNN-based embeddings for improved state-of-the-art (SOTA) fingerprint recognition (both authentication and identification). Our combined architecture, AFR-Net (Attention-Driven Fingerprint Recognition Network), outperforms several baseline transformer and CNN-based models, including a SOTA commercial fingerprint system, Verifinger v12.3, across intra-sensor, cross-sensor, and latent to rolled fingerprint matching datasets. Additionally, we propose a realignment strategy using local embeddings extracted from intermediate feature maps within the networks to refine the global embeddings in low certainty situations, which boosts the overall recognition accuracy significantly across each of the models. This realignment strategy requires no additional training and can be applied as a wrapper to any existing deep learning network (including attention-based, CNN-based, or both) to boost its performance.
\end{abstract}

\begin{IEEEkeywords}
Fingerprint Embeddings, Fingerprint Recognition, Attention, Vision Transformers, Fixed-Length Fingerprint Representations
\end{IEEEkeywords}}

\maketitle

\IEEEdisplaynontitleabstractindextext

%
\IEEEpeerreviewmaketitle

\IEEEraisesectionheading{\section{Introduction}\label{sec:introduction}}

%
%
%
%
\IEEEPARstart{A}{utomated} fingerprint recognition systems have continued to permeate many facets of everyday life, appearing in many civilian and governmental applications over the last several decades~\cite{handbook}. Due to the impressive accuracy of fingerprint recognition algorithms ($0.626\%$ False Non-Match Rate at a False Match Rate of $0.01\%$ on the FVC-ongoing 1:1 hard benchmark~\cite{fvc_ongoing}), researchers have turned their attention to addressing difficult edge-cases where accurate recognition remains challenging, such as partial overlap between two candidate fingerprint images and cross-sensor interoperability (e.g., optical to capacitive, contact to contactless, latent to rolled fingerprints, etc.), as well as other practical problems like template encryption, privacy concerns, and matching latency for large-scale (gallery sizes on the order of tens or hundreds of millions) identification. 

For many reasons, some of which mentioned above (e.g., template encryption and latency), methods for extracting fixed-length fingerprint embeddings using various deep learning approaches have been proposed. Some of these methods were proposed for specific fingerprint-related tasks, such as minutiae extraction~\cite{darlow2017fingerprint, tang2017fingernet} and fingerprint indexing~\cite{index1, index2}, whereas others were aimed at extracting a single ``global" embedding~\cite{learning2019, deepprint, lin2018cnn}. Of these methods, the most common architecture employed is the convolutional neural network (CNN), often utilizing domain knowledge (e.g., minutiae~\cite{deepprint}) and other tricks (e.g., specific loss functions, such as triplet loss~\cite{dong2018triplet}) to improve fingerprint recognition accuracy. More recently, motivated by the success of attention-based Transformers~\cite{vaswani2017attention} in natural language processing, the computer vision field has seen an influx of the use of the vision transformer (ViT) architecture for various computer vision tasks~\cite{dosovitskiy2020image, carion2020end, han2021transformer, liu2021swin}. 

\begin{figure}[!t]
\includegraphics[width=\linewidth]{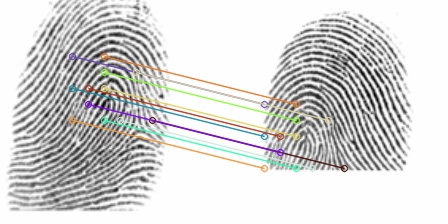} 
\caption{Example correspondence between local features extracted from the intermediate feature maps of our AFR-Net model for two images of the same finger. Note, these local features are not necessarily the same as minutiae points, which are commonly used in fingerprint recognition.}
\label{fig:corr}
\end{figure}

In fact, two studies have already explored the use of ViT for learning discriminative fingerprint embeddings~\cite{tandon2022transformer, grosz2022minutiae}; albeit, with the following limitations: i.) the authors of \cite{tandon2022transformer} supervised their ViT model using a pretrained CNN as a teacher model and thus did not give the transformer architecture the freedom to learn its own representation and ii.) the authors of \cite{grosz2022minutiae} were limited in the data and choice of loss function used to supervise their transformer model, thereby limiting the fingerprint recognition accuracy compared to the baseline ResNet50 model. Nonetheless, the authors in \cite{grosz2022minutiae} did note the complimentary nature between the features learned by the CNN-based ResNet50 model and the attention-based ViT model. This motivated us to evaluate additional attention-based models that bridge the gap between purely CNN and purely attention-based models, in order to leverage the benefits of each. Toward this end, we evaluate two ViT variants (vanilla ViT~\cite{dosovitskiy2020image} and Swin~\cite{liu2021swin}) along with two variants of a CNN model~\cite{he2016deep} (ResNet50 and ResNet101) for fingerprint recognition. In addition, we propose our own architecture, AFR-Net (Attention-Driven Fingerprint Recognition Network), consisting of a shared feature extraction and parallel CNN and attention classification layers.

Even though these models are trained to extract a single, global embedding representing the identity of a given fingerprint image, we make the observation that for both CNN-based and attention-based models, the intermediate feature maps encode local features that are also useful for relating two candidate fingerprint images. Correspondence between these local features can be used to guide the network in placing attention on overlapping regions of the images in order to make a more accurate determination of whether the images are from the same finger. Additionally, these local features are useful in explaining the similarity between two candidate images by directly visualizing the corresponding keypoints, as shown in Figure~\ref{fig:corr}.

More concisely, the contributions of this research are as follows:

\begin{itemize}
    \item Analysis of various attention-based architectures for fingerprint recognition.
    \item Novel architecture for fingerprint recognition, AFR-Net, which incorporates attention layers into the ResNet architecture.
    \item State-of-the-art (SOTA) fingerprint recognition performance (authentication and identification) across several diverse benchmark datasets, including intra-sensor, cross-sensor, contact to contactless, and latent to rolled fingerprint matching.
    \item Novel use of local embeddings extracted from intermediate feature maps to both improve the recognition accuracy and explainability of the model.
    \item Ablation analysis demonstrating the importance of each aspect of our model, including choice of loss function, training dataset size, use of spatial alignment module, and use of local embeddings to refine the global embeddings.
\end{itemize}

\section{Related Work}
Here we briefly discuss the prior literature in deep learning-based fingerprint recognition and the use of vision transformer models for computer vision. For a more in-depth discussion on these topics, refer to one of the many survey papers available (e.g., \cite{minaee2019biometrics} for deep learning in biometrics and \cite{khan2022transformers} for the use of transformers in vision). 

\subsection{Deep Learning for Fingerprint Recognition}
Over the last decade, deep learning has seen a plethora of applications in fingerprint recognition, including minutiae extraction~\cite{darlow2017fingerprint, tang2017fingernet}, fingerprint indexing~\cite{index1, index2}, presentation attack detection~\cite{chugh2018fingerprint, engelsma2019generalizing, uliyan2020anti, grosz2020fingerprint}, synthetic fingerprint generation~\cite{kim2019fingerprint, engelsma2022printsgan, grosz2022spoofgan, wyzykowski2022synthetic}, and fixed-length fingerprint embeddings for recognition~\cite{learning2019, deepprint, lin2018cnn}. For purposes of this paper, we limit our discussion to fixed-length (global) embeddings for fingerprint recognition. 

Among the first studies on extracting global fingerprint embeddings using deep learning was proposed by Li et al.~\cite{learning2019}, which used a fully convolutional neural network to produce a final embedding of 256 dimensions. The authors of \cite{deepprint} then showed improved performance of their fixed-length embedding network by incorporating minutiae domain knowledge as an additional supervision. Similarly, Lin and Kumar incorporated additional fingerprint domain knowledge (minutiae and core point regions) into a multi-Siamese CNN for contact to contactless fingerprint matching~\cite{lin2018cnn}. More recently, \cite{tandon2022transformer} and \cite{grosz2022minutiae} proposed the use of vision transformer architecture for extracting discriminative fixed-length fingerprint embeddings, both showing that incorporating minutiae domain knowledge into ViT improved the performance.

\subsection{Vision Transformers for Biometric Recognition}
Transformers have led to numerous applications across the computer vision field in the past couple of years since they were first introduced for computer vision applications by Doesovitskiy et al. in 2021\cite{dosovitskiy2020image}. The general principle of transformers for computer vision is the use of the attention mechanism for aggregating sets of features across the entire image or within local neighborhoods of the image. Attention was originally introduced in 2015 for sequence modeling by Bahdanau et al.~\cite{bahdanau2014neural} and has been shown to be a useful mechanism in general for operations on a set of features. Today, numerous variants of ViT have been proposed for a wide range of computer vision tasks, including image recognition, generative modeling, multi-model tasks, video processing, low-level vision, etc.~\cite{khan2022transformers}.

Some recent works have explored the use of transformers for biometric recognition across several modalities including face~\cite{zhong2021face}, finger vein~\cite{huang2022fvt}, fingerprint~\cite{tandon2022transformer, grosz2022minutiae}, ear~\cite{alejo2021unconstrained}, gait~\cite{delgado2022exploring}, and keystroke recognition~\cite{stragapede2022mobile}. In this work, we improve upon these previous uses of transformers for fingerprint recognition by evaluating additional attention-based architectures for extracting global fingerprint embeddings.

\section{AFR-Net: Attention-Driven Fingerprint Recognition Model}
Our approach consists of i.) investigating several baseline CNN and attention-based models for fingerprint recognition, ii.) fusing a CNN-based architecture with attention into a single model to leverage the complimentary representations of each, iii.) a strategy to use intermediate local feature maps to refine global embeddings and reduce uncertainty in challenging pairwise fingerprint comparisons, and iv.) use of spatial alignment module to improve recognition performance. Details of each component of our approach are given in the following sections.

\begin{figure*}[!t]
\centering
\includegraphics[width=\linewidth]{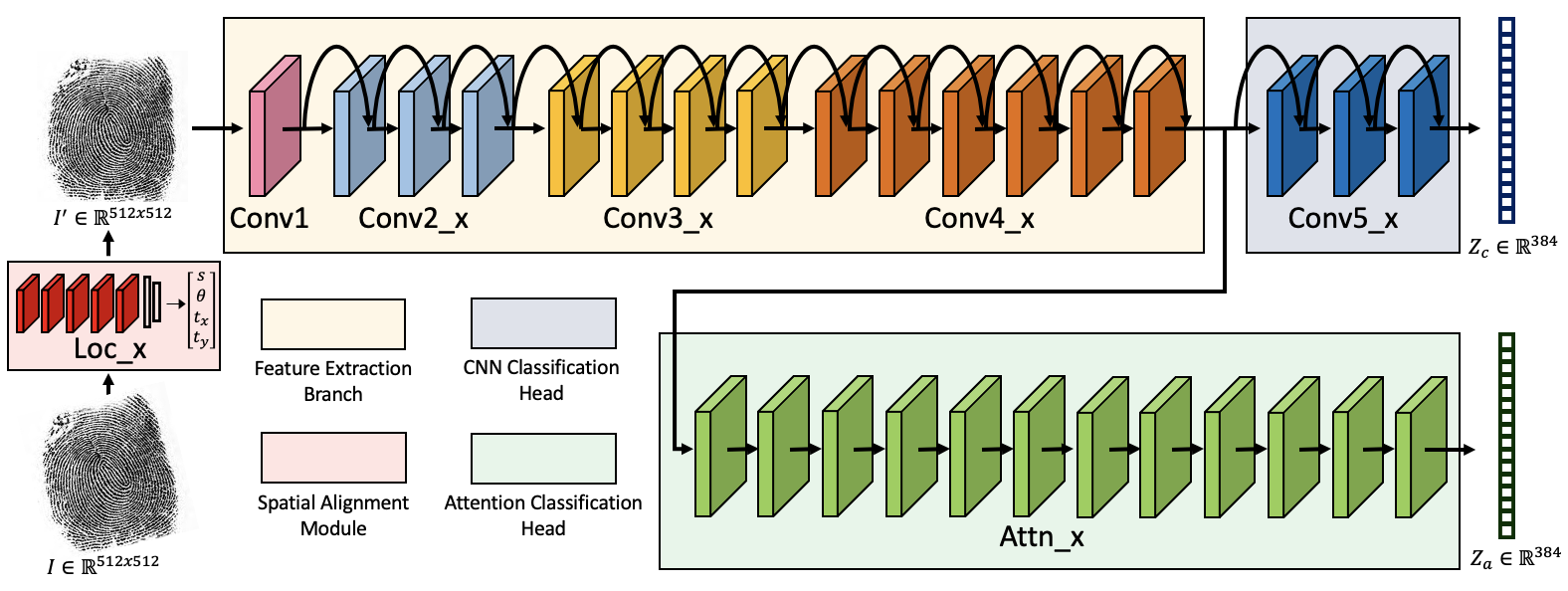} 
\caption{Overview of the AFR-Net architecture. First, input fingerprint images are passed through a spatial alignment module for better alignment of two fingerprints under comparison, then passed through a shared feature extraction, followed by two classification heads (one CNN-based and the other attention-based). For our implementation, we followed the ResNet50 architecture as our backbone and CNN classification head and used 12 multi-headed attention transformer encoder blocks for the attention-based classification head.}
\label{fig:diagram}
\end{figure*}

\subsection{Baseline Methods}
First, we improve on the initial studies (\cite{tandon2022transformer, grosz2022minutiae}) applying ViT to fingerprint recognition to better establish a fair baseline performance of ViT compared to the CNN-based models. This is accomplished by removing the limitations of the previous studies in terms of choice of supervision and size of training dataset used to learn the parameters of the models. We then compare ViT with two variants of the ResNet CNN-based architecture, ResNet50 and ResNet101. For our specific choice of ViT, we decided on the small version with patch size of 16, number of attention heads of 6, and layer depth of 12. We selected this architecture as it presents an adequate trade-off in speed and accuracy compared to other ViT architecture variants.

In addition, we compare the performance of a popular ViT successor, Swin, which utilizes a hierarchical structure and shifted windows for computing attention within local regions of the image. Specifically, we used the small Swin architecture with patch size of 4, window size of 7, and embedding dimension of 96. For another strong baseline comparison, we included the latest version of the SOTA fingerprint recognition system from Neurotechnology, Verifinger v12.3, for our evaluations\footnote{\url{https://neurotechnology.com/verifinger.html}}. According to the FVC On-going competition, Verifinger is the top performing algorithm for the 1:1 fingerprint verification benchmark~\cite{fvc_ongoing}.

\subsection{Proposed AFR-Net Architecture}
Based on previous research suggesting the complimentary nature of ViT and ResNet embeddings, we were motivated to merge the two into a single architecture, referred to as AFR-Net. As shown in Figure~\ref{fig:diagram}, AFR-Net consists of a spatial alignment module, shared CNN feature encoder, CNN classification head, and an attention classification head. The shared alignment module and feature encoder greatly reduces the number of parameters compared to the fusion of the two separate networks and also allows the two classification heads to be trained jointly. The full architectural details of AFR-Net are given in Table~\ref{tab:architecture}.

Due to the two classification heads, we have two bottleneck classification layers which map each of the 384-d embeddings, $Z_c$ and $Z_a$, into a softmax output representing the probability of a sample belonging to one of N classes (identities) in our training dataset. We employ the Additive Angular Margin (ArcFace) loss function to encourage intra-class compactness and inter-class discrepancy of the embeddings of each branch~\cite{deng2019arcface}. Through an ablation study, presented in section~\ref{ablation}, we find that despite the relatively little use of this loss function in previous fingerprint recognition papers~\cite{grosz2022minutiae, ozturk2022minnet}, the ArcFace loss function makes an enormous difference in the performance of our model.

\begin{table}
\caption{AFR-Net architecture details.}
\label{tab:architecture}
\begin{tabularx}{\linewidth}{X || X || M || X}
\noalign{\hrule height 1.5pt}
\textbf{Layer Name} & \textbf{Layer Type} & \textbf{Output Dim.} & \textbf{Parameters} \\
\noalign{\hrule height 1.0pt}
\multicolumn{4}{c}{\textbf{Spatial Alignment Module}} \\
\hline
0. Input & - & 3\times224\times224\\
\hline
1. Loc$_{1}$ & Conv2d & 16\times224\times224 & k=7x7, padding=3\\
\hline
2. Loc$_{2}$ & MaxPool & 16\times112\times112 & k=2x2, stride=2\\
\hline
3. Loc$_{3}$ & Conv2d & 24\times112\times112 & k=5x5, padding=2\\
\hline
4. Loc$_{4}$ & MaxPool & 24\times56\times56 & k=2x2, stride=2\\
\hline
5. Loc$_{5}$ & Conv2d & 32\times56\times56 & k=3x3, padding=1\\
\hline
6. Loc$_{6}$ & MaxPool & 32\times28\times28 & k=2x2, stride=2\\
\hline
7. Loc$_{7}$ & Conv2d & 48\times28\times28 & k=3x3, padding=1\\
\hline
8. Loc$_{8}$ & MaxPool & 48\times14\times14 & k=2x2, stride=2\\
\hline
9. Loc$_{9}$ & Conv2d & 64\times14\times14 & k=3x3, padding=1\\
\hline
10. Loc$_{10}$ & MaxPool & 64\times7\times7 & k=2x2, stride=2\\
\hline
11. Loc$_{11}$ & Linear & 32 & bias=True \\
\hline
12. Loc$_{12}$ & Linear & 4 & bias=True \\
\noalign{\hrule height 1.0pt}
\multicolumn{4}{c}{\textbf{Feature Extraction Branch}} \\
\hline
13. Conv$_{1}$ & Conv2d & 64\times112\times112 & k=7x7, padding=3, stride=2\\
\hline
14. Conv$_{2}$ & Conv2d & 256\times56\times56 &
    $\begin{bmatrix}
        1\times1, 64 \\
        3\times3, 64 \\
        1\times1, 256
    \end{bmatrix}$x3\\
\hline
15. Conv$_{3}$ & Conv2d & 512\times28\times28 &
    $\begin{bmatrix}
        1\times1, 128 \\
        3\times3, 128 \\
        1\times1, 512
    \end{bmatrix}$x4\\
\hline
16. Conv$_{4}$ & Conv2d & 1024\times14\times14 &
    $\begin{bmatrix}
        1\times1, 256 \\
        3\times3, 256 \\
        1\times1, 1024
    \end{bmatrix}$x6\\
\noalign{\hrule height 1.0pt}
\multicolumn{4}{c}{\textbf{CNN Classification Head}} \\
\hline
17. Conv$_{5}$ & Conv2d & 2048\times7\times7 &
    $\begin{bmatrix}
        1\times1, 512 \\
        3\times3, 512 \\
        1\times1, 2048
    \end{bmatrix}$x3\\
\hline
18. Z$_{c}$ & Linear & 384 & bias=True \\
\noalign{\hrule height 1.0pt}
\multicolumn{4}{c}{\textbf{Attention Classification Head}} \\
\hline
19. Embed & MLP & 384\times14\times14 & in=1024, hid=1024, out=384 \\
\hline
\specialcell{20. Pos Embed} & Concat & 384\times197 & - \\
\hline
21. Attn$_{1-12}$ & Self-Attention + MLP & 384\times197 & in=384, hid=1536, out=384 \\
\hline
22. Z$_{a}$ & Linear & 384 & bias=True \\
\noalign{\hrule height 1.5pt}
\end{tabularx}
\end{table}

\subsection{Global Embedding Refinement via Local Embeddings}
As noted in the introduction, and demonstrated in Figure~\ref{fig:corr}, we find that the intermediate feature maps of our AFR-Net model (and in general, all the deep learning models evaluated in this work) encode local descriptors (i.e., embeddings) of the input images. These local descriptors can be matched between two fingerprint images and used to compute a correspondence between similar regions. Given the surprising accuracy of these local embeddings in locating corresponding points of interest between two images, we devise a strategy to use these corresponding regions of interest as a sort of hard attention for the model to refine the global embeddings based on just the overlapping regions present in both images.

\begin{figure}
\centering
\includegraphics[width=\linewidth]{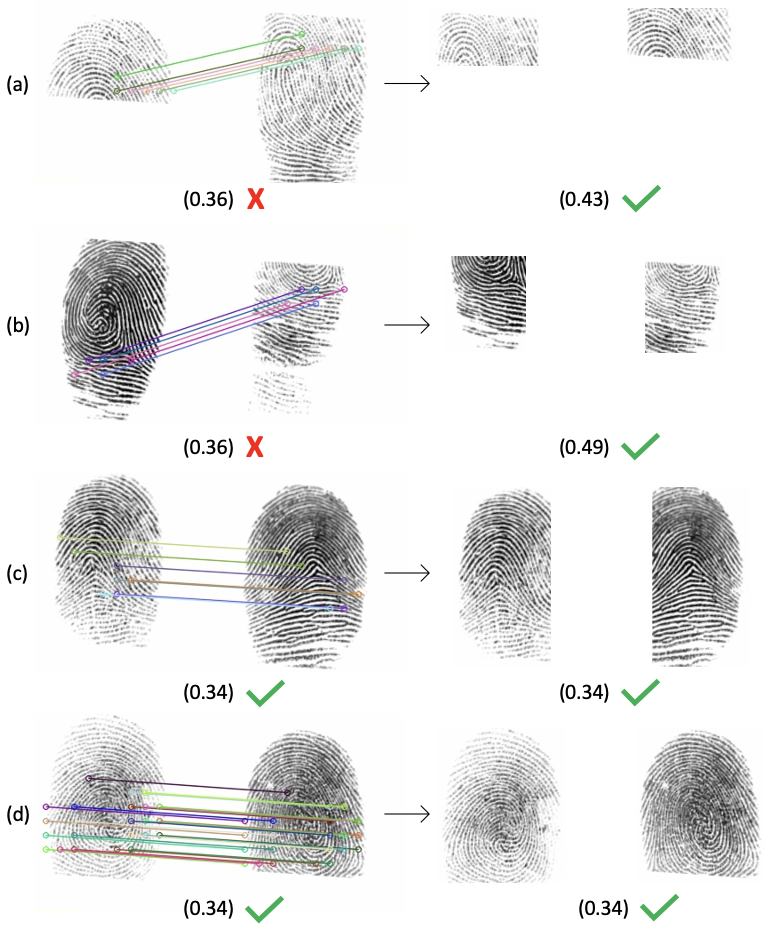} 
\caption{Example genuine (a and b) and imposter (c and d) pairs from FVC 2002 DB1A before and after realignment, with corresponding similarity scores from AFR-Net. Both genuine scores are pushed above the FAR=0.1\% threshold of 0.36, whereas both imposter scores remained below the threshold.}
\label{fig:example_crops}
\end{figure}

Some examples of this process are demonstrated in Figure~\ref{fig:example_crops}, where the correspondence between local embeddings is used to compute an affine transformation between the image pairs. Then, the non-overlapping fingerprint regions are masked and each image is presented to the network for a second time to yield a new set of embeddings. Finally, a second similarity score between the masked images is computed via a cosine similarity between the new embeddings. The similarity between the masked regions is combined via a weighted sum with the similarity score obtained from the original images to obtain a final similarity score. 

For ResNet50, ResNet101, and AFR-Net, we take the last output of the Conv4 layer as our local embeddings, which has dimensions of 14x14x1024. For ViT and Swin, we take the final patch embeddings at the output of the last attention layer as the local embeddings, which has dimensions of 14x14x384. In all cases, each of these 196 local descriptors corresponds to a single 16x16 patch of the input fingerprint image. We assign the center of each patch as the keypoint associated with the corresponding local embedding when computing the correspondence points between two fingerprint images. 

\begin{figure*}[!t]
\centering
\includegraphics[width=\linewidth]{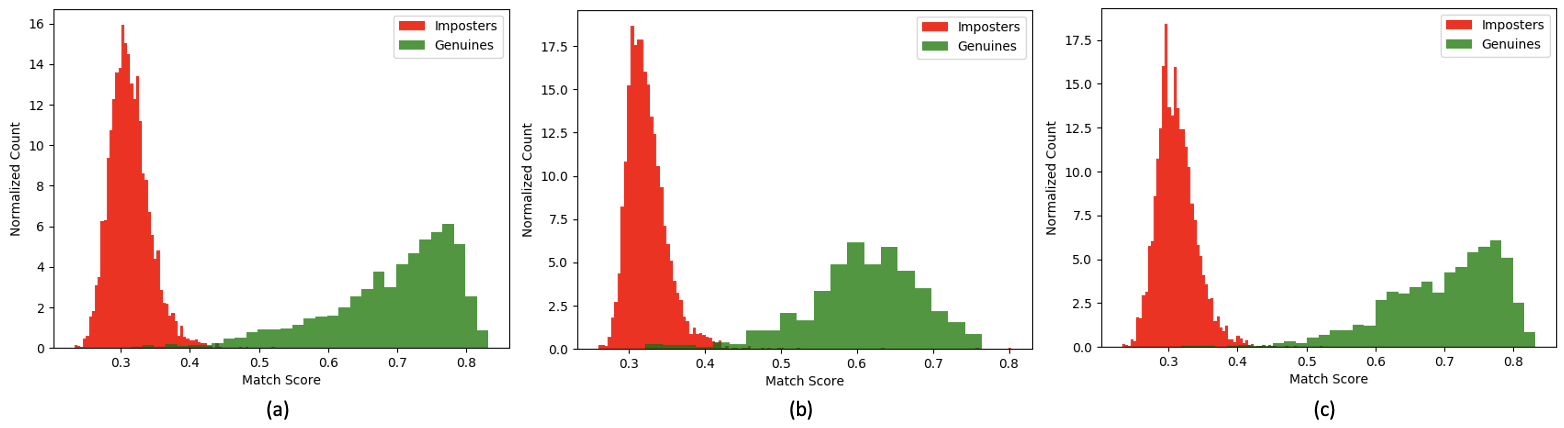} 
\caption{Similarity score distributions for (a) original image embeddings, (b) embeddings after refinement, and (c) weighted average of original and refined embeddings. Similarity scores are computed with the AFR-Net model on the FVC 2002 DB3A dataset, where the TAR @ 0.1\% FAR for the original embeddings is 98.43\%, 91.32\% for the refined embeddings, and 99.36\% after the weighted score fusion.}
\label{fig:histograms}
\end{figure*}

Indeed, computing the correspondence between sets of local descriptors of two images is time consuming, especially if computing a brute force exhaustive search to establish a 1:1 correspondence between matched descriptors. For this reason, we only employ the re-weighting strategy in low certainty scenarios (when the similarity score is close to the match threshold) to keep the amortized latency of our algorithm approximately the same as without the re-weighting process; that is, we only utilize the local descriptors if the similarity score between the original global embeddings falls between a specified range [$s_l$, $s_h$]. Values of 0.3 and 0.6 for $s_l$ and $s_h$, respectively, were empirically determined on our validation dataset to work well across all our models. Furthermore, if a valid homography computed between corresponding local regions cannot be obtained (e.g., if the scale, rotation, and/or translation parameters exceed expected limits), we fall back to the original similarity score as to not further degrade the comparison by computing a new set of embeddings from images which have been corrupted due to poorly behaved transformation matrices. Figure~\ref{fig:histograms}, shows the genuine and imposter score distributions for our AFR-Net model on the FVC 2002 DB3A dataset, where we experienced the biggest increase in performance after re-weighting the predictions using this method. In figure~\ref{fig:histograms}, we show (a) the original score distributions, (b) the scores computed on the refined embeddings, and (c) the fused score distributions after the weighted averaging. The full algorithm of this process is detailed in Algorithm~\ref{alg:resattn}.

\begin{algorithm}
\caption{Compute similarity between input fingerprint pairs with AFR-Net.}\label{alg:resattn}
\begin{algorithmic}[1]
\Procedure{Match}{$I_1,I_2$}
\State $w_1 := 0.2$
\State $w_2 \gets 1 - w_1$
\State $w_3 := 0.5$
\State $w_4 \gets 1 - w_3$
\State $s_l, s_h := [0.3,  0.6]$
\State

\State $Z_c^1, Z_a^1, L_1 \gets AFRnet(I_1)$
\State $Z_c^2, Z_a^2, L_2 \gets AFRnet(I_2)$
\State

\State $s \gets w_1(Z_{c1} \cdot Z_{c2}) + w_2(Z_{a1} \cdot Z_{a2})$
\State

\If{$s_l \leq s \leq s_h$}
    \State $kp1, kp2 \gets getCorr(I_1, I_2, L_1, L_2)$
    \State $M \gets getHomography(kp1, kp2)$
    \If{$homographyOK(M)$}
        \State $I_1' \gets M I_1$ 
        \State $C_1, C_2 \gets cropOverlap(I_1', I_2)$
        
        \State $Z_c^1, Z_a^1, \_ \gets AFRnet(C_1)$
        \State $Z_c^2, Z_a^2, \_ \gets AFRnet(C_2)$

        \State $s' \gets w_1(Z_c^1 \cdot Z_c^2) + w_2(Z_a^1 \cdot Z_a^2)$

        \State $s \gets w_3s + w_4s'$        
    \EndIf
\EndIf
\State

\State \textbf{return} $s$
\EndProcedure
\end{algorithmic}
\end{algorithm}

There are likely more sophisticated, faster alternatives to a brute force algorithm for computing the correspondence between sets of local embeddings, as well as better ways of aggregating the similarity scores between matched local embeddings themselves. However, we leave those areas of exploration for future work as the current algorithm seems to improve the results significantly across all models and datasets and has the nice interpretation of giving the network a second glance at regions of interest in uncertain cases - much like a human fingerprint examiner would. Nonetheless, some suggestions for future extensions would include the use of a graphical neural network (GNN) or attention mechanism to more intelligently aggregate the sets of local descriptors between two images.

\subsection{Spatial Alignment Module}
As has been noted in previous literature on fingerprint recognition~\cite{deepprint, grosz2021c2cl, tang2022end, he2020partial, dabouei2019deep}, Spatial Transformer Networks~\cite{jaderberg2015spatial} have been shown to be highly effective in aligning input fingerprints for improved recognition accuracy across a wide range of tasks (e.g., contact to contactless fingerprint matching, partial fingerprint recognition, etc.). That, coupled with our observation that the local descriptors used in our realignment procedure are not rotation invariant, we were motivated to include a spatial alignment module into the architecture of our AFR-Net model and each of the baseline models. The details of our spatial alignment module are given in Table~\ref{tab:architecture}, along with the details of the rest of the AFR-Net network architecture. Lastly, in the ablation portion of our experimental section, we further emphasize the benefit of incorporating the spatial alignment module into our fingerprint recognition network.

\subsection{Training Details}
AFR-Net and all baseline models, excluding Verifinger, were trained with an ArcFace loss function with a margin of 0.5, learning rate of 1e-4, weight decay of 2e-5, and polynomial learning rate decay function with a power of 3 and minimum learning rate of 1e-5. The AFR-Net, ResNet100, and Swin models were trained with a batch size of 64 across four Nvidia GeForce RTX 2080 Ti GPUs, whereas the ResNet50 and ViT models were trained with a batch size of 128. AFR-Net, ResNet50, and ResNet100 were trained with the Adam optimizer~\cite{kingma2014adam} and ViT and Swin were trained with the AdamW optimizer~\cite{loshchilov2017decoupled}. The maximum number of epochs for all models was set to 75; however, the number of epochs trained for the final saved models varied based on the highest validation accuracy computed during training on a hold-out validation dataset. Finally, we initialize each model using the pre-trained ImageNet weights made available by the open-sourced pytorch-image-models git repository~\cite{rw2019timm}. 

\section{Experimental Results}
In this section, we discuss the training and evaluation datasets used, the authentication and identification results achieved by our method in comparison with the baseline methods, latency and performance trade-off between the methods, and an ablation analysis to highlight the contributions of individual components of our algorithm.

\begin{figure*}
\centering
\includegraphics[height=\textheight]{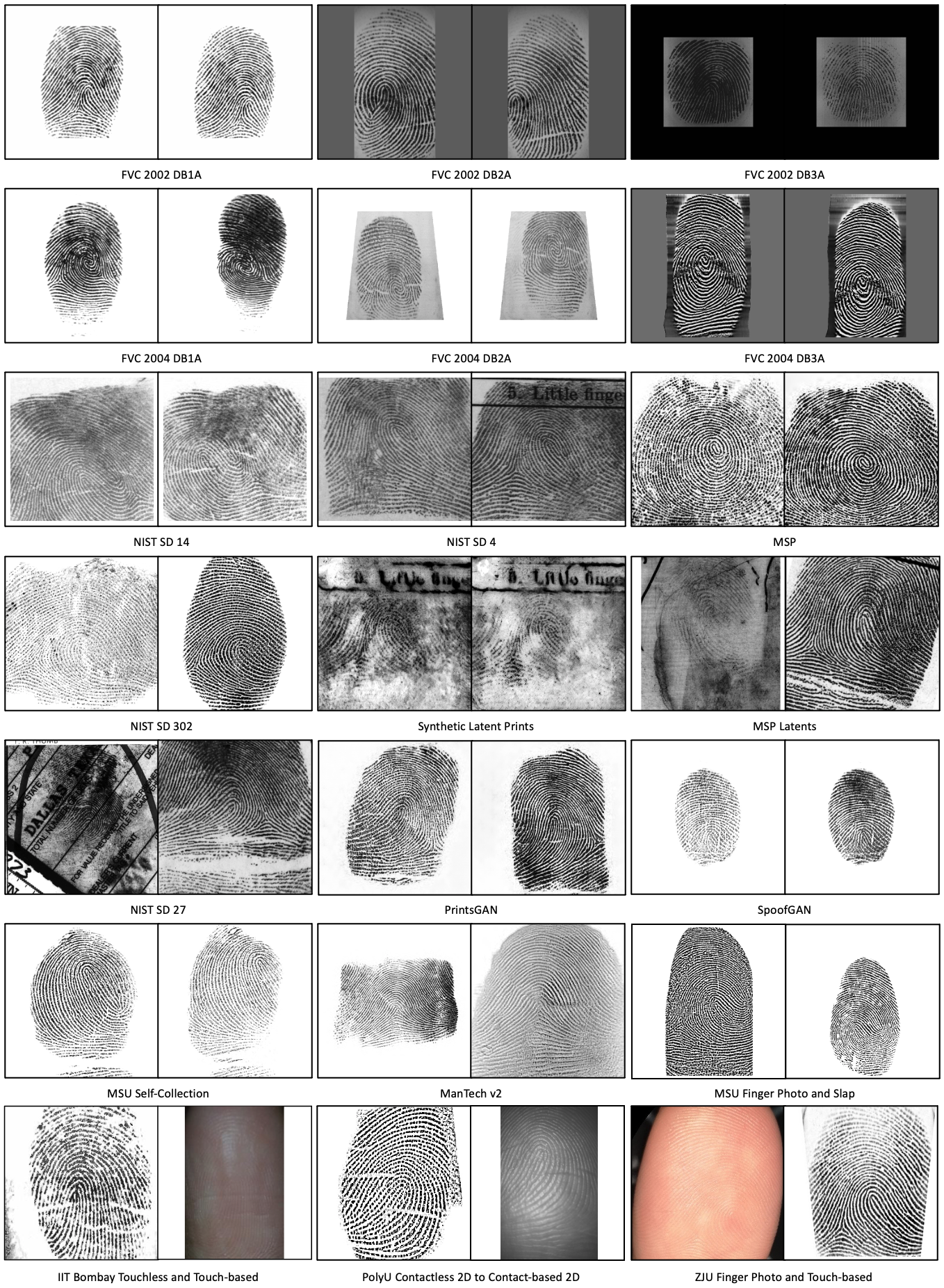} 
\caption{Example image pairs from each of the datasets used in this paper. See Table~\ref{tab:datasets} for details.}
\label{fig:datasets}
\end{figure*}

\subsection{Datasets}
For training our models, we aggregate a large number of fingerprint recognition datasets with diverse characteristics ranging from rolled fingerprints~\cite{yoon2015longitudinal, sd4}, plain (i.e., slap) fingerprints, mixture of rolled and plain fingerprints~\cite{nist302}, contactless (e.g., from mobile phone cameras) fingerprints~\cite{ericson2015evaluation, deb2018matching, birajadar2019towards}, latent fingerprints (from the Michigan State Police (MSP) Latent Database), and even recently released synthetic fingerprints~\cite{engelsma2022printsgan, grosz2022spoofgan, wyzykowski2022synthetic}. Example images from each of these datasets are shown in Figure~\ref{fig:datasets}. A small portion of the total training dataset was reserved for validation. In total, our aggregated training dataset contains 1.3M images for training and 3,814 images for validation. Further information regarding the number of unique fingers, images per dataset, and fingerprint type is given in Table~\ref{tab:datasets}.

Our evaluation datasets are just as diverse as our training datasets and include challenging scenarios such as contact to contactless fingerprint matching~\cite{lin2018matching, grosz2021c2cl}, varying sensor types for both rolled and slap prints (e.g., optical, capacitive, thermal swipe, etc.)~\cite{sd14, fvc2002, fvc2004}, latent to rolled fingerprint matching~\cite{sd27}, and even rolled to plain fingerprint matching (as is the case in the NIST SD 302 dataset~\cite{nist302}\footnote{We have reserved 200 of the 2,000 unique fingers in the NIST SD 302 for testing; these 200 fingers are completely disjoint from the fingers used in our training and validation partitions.}). 

\begin{table}
\centering
\caption{Fingerprint Datasets used in this study.}
\label{tab:datasets}
\begin{tabular}{lcc}
\noalign{\hrule height 1.5pt}
\textbf{Train Dataset}      & \textbf{\# Fingers} & \textbf{\# images} \\
\noalign{\hrule height 1.0pt}
MSP$^\dagger$~\cite{yoon2015longitudinal}                         & 37,411               & 447,988             \\
\hline
NIST SD 302~\cite{nist302}                 & 1,600                & 20,008              \\
\hline
MSU Self-Collection$^\dagger$                    & 4,582                & 57,813              \\
\hline
PrintsGAN~\cite{engelsma2022printsgan}                   & 34,985               & 524,775             \\
\hline
SpoofGAN~\cite{grosz2022spoofgan}                    & 10,000               & 150,000             \\
\hline
\leftcell{MSU Finger Photo\\and Slap Database}~\cite{deb2018matching}                      & 1,243                & 5,220               \\
\hline
\leftcell{IIT Bombay Touchless\\and Touch-based Database}~\cite{birajadar2019towards}                  & 200                  & 1,600               \\
\hline
ManTech Phase 2~\cite{ericson2015evaluation}                    & 4,535                & 64,061              \\
\hline
Synthetic Latent Prints~\cite{wyzykowski2022synthetic}     & 2,000                & 16,000              \\
\hline
NIST SD 4$^\dagger$~\cite{sd4}                   & 2,000                & 4,000               \\
\noalign{\hrule height 1.0pt}
\textbf{Validation Dataset} & \textbf{\# Fingers} & \textbf{\# Images} \\
\noalign{\hrule height 1.0pt}
\leftcell{MSU Finger Photo\\and Slap Database}~\cite{deb2018matching}                      & 110                  & 200                 \\
\hline
MSP Latent$^\dagger$~\cite{yoon2015longitudinal}                  & 524                  & 1086                \\
\hline
NIST SD 302~\cite{nist302}                 & 200                  & 2528                \\
\hline
\noalign{\hrule height 1.0pt}
\textbf{Test Dataset}       & \textbf{\# Fingers} & \textbf{\# Images} \\
\noalign{\hrule height 1.0pt}
FVC 2002 DB1A~\cite{fvc2002}               & 100                  & 800                 \\
\hline
FVC 2002 DB2A~\cite{fvc2002}               & 100                  & 800                 \\
\hline
FVC 2002 DB3A~\cite{fvc2002}               & 100                  & 800                 \\
\hline
FVC 2004 DB1A~\cite{fvc2004}               & 100                  & 800                 \\
\hline
FVC 2004 DB2A~\cite{fvc2004}               & 100                  & 800                 \\
\hline
FVC 2004 DB3A~\cite{fvc2004}               & 100                  & 800                 \\
\hline
NIST SD 14$^\dagger$~\cite{sd14}                  & 2700                 & 5,400               \\
\hline
NIST SD 302~\cite{nist302}                 & 200                  & 2,548               \\
\hline
NIST SD 27$^\dagger$~\cite{sd27}                  & 258                  & 516                 \\
\hline
\leftcell{PolyU Contactless 2D to\\Contact-based 2D Database}~\cite{lin2018matching}                      & 160                  & 960                 \\
\hline
\leftcell{ZJU Finger Photo and\\Touch-based Database}~\cite{grosz2021c2cl}                         & 824                  & 19,776              \\
\noalign{\hrule height 1.5pt}
\multicolumn{3}{p{0.95\linewidth}}{$^\dagger$ Not publicly available. NIST SD 4, NIST SD 14 and NIST SD 27 were publicly available but were later removed from public domain by NIST. MSP and MSP Latent databases are operational forensic datasets which cannot be released for privacy reasons and per our NDA with the Michigan State Police (MSP).}\\
\end{tabular}
\end{table}

\subsection{Authentication Results}
We report authentication performance of our method across 11 different evaluation datasets of varying characteristics. The results are given in Table~\ref{tab:authentication} as the true accept rate (TAR) at a false accept rate (FAR) of 0.01\% (FAR=0.1\% in the case of the FVC datasets in order to follow the established protocols) and the full Receiver Operating Characteristic (ROC) curves are given in the appendix. Besides for the established protocol on the FVC datasets, we compute all possible genuine and imposter pairs for our evaluations.

According to the results in Table~\ref{tab:architecture}, AFR-Net outperforms the baseline methods on 9 out of the 11 datasets and shows competitive performance on the two datasets where it comes in second place (99.96\% vs. 100\% and 99.36\% vs. 99.54\% for FVC 2002 DB2A and DB3A, respectively). We show especially impressive performance in cross-sensor (TAR=96.11\% on NIST SD 302) and contact to contactless matching (TAR=98.73\% and TAR=98.70\% on PolyU and ZJU datasets, respectively), as well as latent to rolled fingerprint matching on the challenging NIST SD 27 dataset, where our method out performs Verifinger v12.3 (TAR=63.18\% to TAR=61.63\%). 

AFR-Net, and even our baseline ResNet and ViT variants, show substantial improvement over previous fixed-length, global representation networks for fingerprint recognition. For example, DeepPrint, one of the top performing models in the open literature, achieves a TAR of 97.53\% and 98.55\% on FVC 2004 DB1A and NIST SD 14, respectively. However, given the older architectures (Inception v4 in the case of DeepPrint), loss functions, and smaller training datasets, this increase in performance is not all that surprising. In fact, this is why we benchmarked our AFR-Net model against several more recent architectures like ResNet, ViT, and their variants; all of which were trained and evaluated on the same datasets, allowing for a more fair comparison with AFR-Net.

For all the methods, we show improved performance with using the local embeddings to realign the images as a way to refine the global embeddings and improve the resulting similarity scores. The performance improvement was most pronounced for datasets with frequent partial fingerprints, such as FVC 2002 DB3A and DB1A. For example, the average performance across all the methods on FVC 2002 DB3A improved from 94.46\% to 96.26\%, a 32.5\% reduction in error. Intuitively, this realignment process has the effect of slightly improving the similarity scores between borderline genuine fingerprint pairs, by forcing the network to focus on overlapping regions in the images, and does not appreciably effect the borderline imposter scores. Thereby, pushing some of the borderline genuine matches above the rejection threshold. 

If comparing just the CNN-based models (ResNet50 and ResNet101) vs. the attention-based models (ViT and Swin), the performance in terms of matching accuracy is quite comparable; however, in terms of number of parameters, ViT and Swin have substantially smaller footprints. As a result, the training time to reach convergence of these networks was significantly faster than the ResNet models, especially for ViT due to it's low latency as well (which is comparable to ResNet50). Finally, for the most part, Swin outperformed ViT in terms of accuracy across many of the datasets, but it does have more than twice the parameters and 3 times the latency of ViT, making it perhaps not as preferable in some situations.

\begin{table*}
\caption{Authentication (1:1 matching) results.}
\label{tab:authentication}
\begin{tabular}{c|c|c|ccc|ccc|ccccc}
\noalign{\hrule height 1.5pt}
\multirow{3}{*}{\textbf{Model}} & \multirow{3}{*}{\specialcell{\textbf{\# Params.}\\\textbf{(M)}}} & \multirow{3}{*}{\specialcell{\textbf{Inference}\\\textbf{Speed}$^\ddagger$\\\textbf{(ms)}}} & \multicolumn{6}{c|}{\textbf{TAR (\%) @ FAR=0.1\%}$^*$}                                                                        & \multicolumn{5}{c}{\textbf{TAR (\%) @ FAR=0.01\%}}                                                                                                \\
\cline{4-14}
                  &                                     &                                                               & \multicolumn{3}{c|}{\textbf{FVC 2002}}                    & \multicolumn{3}{c|}{\textbf{FVC 2004}}                                   & \multirow{2}{*}{\specialcell{\textbf{NIST}\\\textbf{SD14}}} & \multirow{2}{*}{\specialcell{\textbf{NIST}\\\textbf{SD302}}} & \multirow{2}{*}{\specialcell{\textbf{NIST}\\\textbf{SD27}}} & \multirow{2}{*}{\textbf{PolyU}} & \multirow{2}{*}{\textbf{ZJU}} \\
                  &                                     &                                                               & \textbf{DB1A}          & \textbf{DB2A}  & \textbf{DB3A} &                \textbf{DB1A}             &           \textbf{DB2A}                   &           \textbf{DB3A}                  &                        &          &  &        &      \\
\noalign{\hrule height 1.0pt}
\specialcell{Verifinger\\v12.3}  & N/A                                 &             600                                                   & 99.96         & 99.86          & \textbf{99.54} & 98.79          &      98.46                 &           99.86            & \textbf{99.93}              & 93.26                        & 61.63                       & 95.39                  & 96.88                \\
\noalign{\hrule height 1.0pt}
ResNet50          &          62.21                           &         4.34                                                       & 99.50 & 99.93          & 93.96          & 99.96          & 99.36        & \textbf{100}          & \textbf{99.93}              & 94.70                         & 53.88                       & 96.94                  & 98.28                \\
ResNet101         &                81.20                     &                 7.58                                      & 99.57         & 99.61 & 94.5           & 99.96 & 99.29        & 99.82                 & \textbf{99.93}                       & 93.79                        & 56.59                       & 96.48                  & 97.71                \\
ViT               &            \textbf{21.83}                         &                      \textbf{4.12}                                          & 99.29         & 99.68 & 93.00             & 99.86 & 98.89                 & 99.50                  & 99.67                       & 93.49                        & 46.51                       & 92.38                  & 98.08                \\
Swin              &                  52.69                   &                11.66                                                & 99.75         & 99.79          & 92.43          & 99.75          & 99.04                 & 99.82                 & 99.89                       & 95.46                        & 44.96                       & 96.15                  & 98.33                \\
AFR-Net          &                  85.02                   &                          8.42                                     & 99.86         & 99.96          & 98.43          & \textbf{100}   & 99.36        & \textbf{100}          & \textbf{99.93}              & 95.46                        & \textbf{63.18}              & 98.23                  & 98.68                \\
\noalign{\hrule height 1.0pt}
ResNet50$^\dagger$        &              62.21              & 10.37                                                         & 99.86         & \textbf{100}   & 95.54 & \textbf{100}   & 99.32        & \textbf{100}          & \textbf{99.93}              & 95.40                & 53.88              & 97.60                   & 98.04                \\
ResNet101$^\dagger$       &            81.20                         &              14.19                                                  & 99.75         & 99.75          & 96.11          & \textbf{100}   & 99.34                 & \textbf{100}          & 99.89                       & 94.62                        & 56.59                       & 97.15                  & 97.78                \\
ViT$^\dagger$             &              \textbf{21.83}                       &                 10.11                                               & 99.54         & 99.75          & 95.04          & 99.93          & 99.21                 & 99.82                 & 99.74                       & 94.25                        & 46.51                       & 94.08                  & 98.29                \\
Swin$^\dagger$            &                52.69                     &                 19.00                                               & 99.86         & 99.82          & 95.25          & 99.82          & 99.07                 & 99.89                 & 99.89              & 95.70                         & 44.96                       & 95.95                  & 98.07                \\
AFR-Net$^\dagger$        &            85.02                         &                   15.18                                            & \textbf{100}  & 99.96          & 99.36          & \textbf{100}   & \textbf{99.64}        & \textbf{100}          & \textbf{99.93}              & \textbf{96.11}               & \textbf{63.18}              & \textbf{98.73}         & \textbf{98.70}       \\
\noalign{\hrule height 1.5pt}
\multicolumn{14}{p{0.95\linewidth}}{$^\ddagger$ Computed on an Nvidia GeForce RTX 2080 Ti.}\\
\multicolumn{14}{p{0.95\linewidth}}{$^*$ Following the FVC protocol of 2,800 genuine pairs and 4,950 imposter pairs.}\\
\multicolumn{14}{p{0.95\linewidth}}{$^\dagger$ With re-weighting using local embeddings.}\\
\end{tabular}
\end{table*}

\subsection{Identification Results}
\begin{figure*}
\centering
\includegraphics[width=\linewidth]{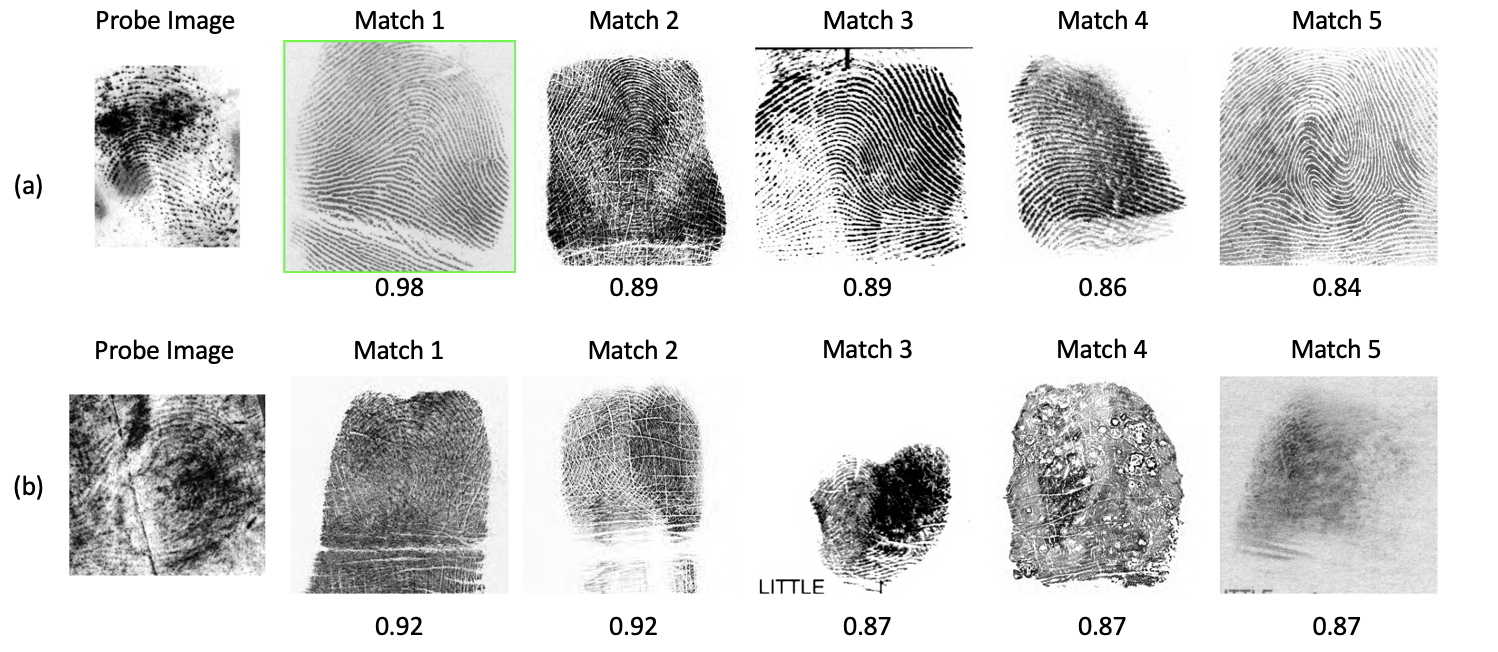} 
\caption{Example (a) successful and (b) unsuccessful search results for two NIST SD 27 latent probe fingerprints. In the successful case, the latent probe image is matched with its corresponding mate with a similarity score of 0.98; whereas, in the unsuccessful case, the poor quality latent image produces high similarity scores with other low quality images in the gallery and is not matched with the correct mate until rank 10 (with a similarity score of 0.84).}
\label{fig:search_examples}
\end{figure*}

\begin{figure}
\centering
\includegraphics[width=\linewidth]{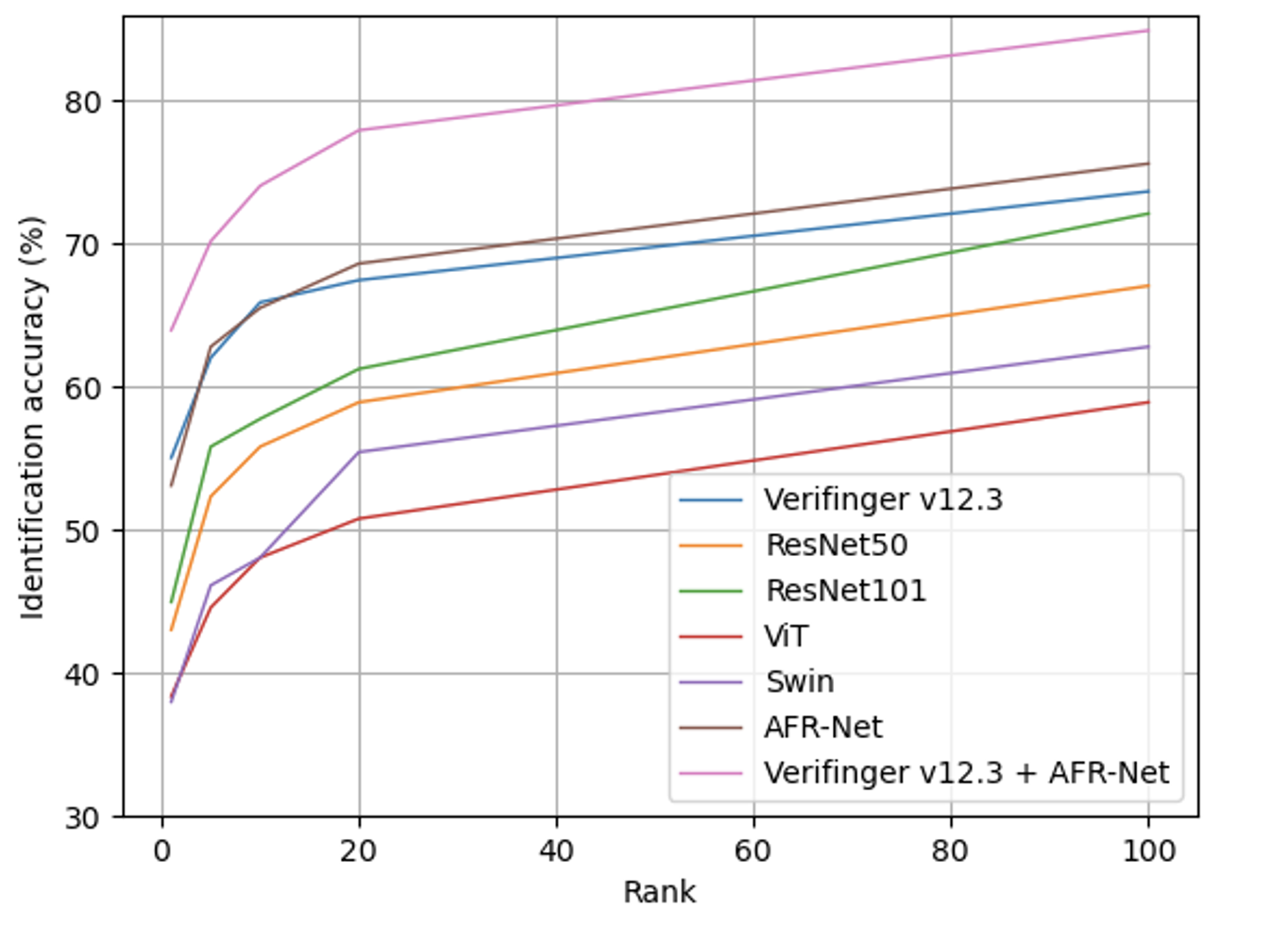} 
\caption{Cumulative match characteristic (CMC) curve with NIST SD 27 latent probes and a gallery of 100K rolled fingerprints plus the NIST SD 27 mated rolled fingerprint pairs.}
\label{fig:cmc}
\end{figure}

We used the NIST SD 27 latent fingerprint dataset and a gallery of 100K rolled fingerprints from the MSP fingerprint dataset to evaluate the closed-set identification (i.e., 1:N search) performance of our models\footnote{These 100K images are completely disjoint from the 448K fingerprint images from MSP used for training.}. According to the cumulative match characteristic (CMC) curve shown in Figure~\ref{fig:cmc} and the identification performance at specific retrieval ranks given in Table~\ref{fig:cmc}, AFR-Net is competitive with Verifinger v12.3 and outperforms all the rest of the baseline methods by a substantial margin. The rank-1 accuracy of Verifinger is 55.04\%, compared to 53.10\% with AFR-Net but AFR-Net surpasses Verifinger at higher retrieval ranks. The next closest performing model was ResNet101, with a rank-1 accuracy of 44.96\%. Some examples image retrievals when (a) the correct mate was returned at rank-1 and (b) when the correct mate was not returned in the top five images are shown in Figure~\ref{fig:search_examples}. In the successful case, the latent probe image is of relatively high quality and is able to match with it's corresponding mate with a similarity score far above the other returned matches; however, in the failure case, the latent image is of very poor quality and returns high similarity scores with other poor quality images in the gallery.

Despite impressive performance of our model compared to the baseline methods, we should note that latent fingerprint identification is a challenging task that requires targeted segmentation, enhancement, and matching strategies to achieve SOTA performance, as is demonstrated in these prior latent identification studies~\cite{jain2010latent, cao2018automated, cao2019end, ozturk2022minnet}. For our evaluation, we only used manual bounding box annotations to locate the latent fingerprints prior to matching, but we did not use any other preprocessing or enhancement; thus, our performance could be further improved for latent to rolled fingerprint matching. Additionally, since we do not use minutiae or any other fingerprint domain knowledge in designing AFR-Net, our model may be at a disadvantage compared to the SOTA latent matchers, since minutiae have been shown to be a useful feature for matching very low quality latents~\cite{cao2019end}. Nonetheless, AFR-Net still performs reasonably well compared to Verifinger, which is also not intended for latent to rolled fingerprint matching but we can assume does incorporate some fingerprint domain knowledge (enhancement, minutiae, etc.). 

Furthermore, we observed that the fusion of the two matchers (Verifinger v12.3 plus AFR-Net) leads to a significant boost in retrieval accuracy (rank-1 accuracy of about 64\% compared to 55.04\% for Verifinger and 53.10\% for AFR-Net). Still, there is room for improvement as the SOTA rank-1 retrieval rate for NIST SD 27 against a gallery of 100K rolled fingerprint is 65.7\%, according to Cao et al.~\cite{cao2019end}. We also evaluated the fusion of ResNet50 and ViT, which performed worse compared to using just AFR-Net (rank 1 retrieval rate of 49.61\% vs. 53.10\%). Thus, not only does incorporating both architectures into one save on latency and model size, as is done in AFR-Net, it also leads to better fingerprint recognition performance over the fusion of both individual models. 

Lastly, we evaluated our model's performance for rolled to rolled fingerprint search using NIST SD 14. Consistent with previous studies~\cite{deepprint}, we used the last 2,700 images from NIST SD 14 as probes and their corresponding mates with the same 100K rolled images from MSP as the gallery. AFR-Net achieves a rank 1 retrieval rate of 99.78\%, which is an improvement over the previous SOTA performance of 99.20\% by DeepPrint~\cite{deepprint}.

\begin{table}
\caption{Closed-set identification performance on NIST SD 27 (rolled to latent comparison) at varying retrieval ranks (\%) against a gallery of 100K rolled fingerprints.}
\label{tab:cmc}
\begin{tabular}{cccccc}
\noalign{\hrule height 1.5pt}
\multicolumn{1}{c}{\textbf{Model}} & \textbf{Rank 1} & \textbf{Rank 5} & \textbf{Rank 10} & \textbf{Rank 20} & \textbf{Rank 100} \\
\noalign{\hrule height 1.0pt}
\specialcell{Verifinger\\v12.3} & \textbf{55.04} & 62.02 & \textbf{65.89} & 67.44 & 73.64 \\
\noalign{\hrule height 1.0pt}
ResNet50             & 43.02 & 52.33 & 55.81 & 58.91 & 67.05 \\
\hline
ResNet101            & 44.96 & 55.81 & 57.75 & 61.24 & 72.09 \\
\hline
ViT                  & 38.37 & 44.57 & 48.06 & 50.78 & 58.91 \\
\hline
Swin                 & 37.98 & 46.12 & 48.06 & 55.43 & 62.79 \\
\hline
AFR-Net             & 53.10 & \textbf{62.79} & 65.50 & \textbf{68.6} & \textbf{75.58} \\
\noalign{\hrule height 1.5pt}
\end{tabular}
\end{table}


\subsection{Latency}
The inference speed of each method is given in Table~\ref{tab:authentication}, along with the number of parameters of each network. Of the models that we compared, the one with the least number of parameters is ViT (21.83M), followed by Swin (52.69M) and ResNet50 (62.21M). ViT also has the lowest latency of 4.12ms, followed closely by ResNet50 with a latency of 4.34ms. AFR-Net has 85.02M parameters, roughly equivalent to the number of parameters as ResNet50 and ViT combined, but is still comparable to the number of parameters as ResNet101 (81.20M).

In terms of performance vs. latency trade-off, ResNet50 outperformed ResNet101 on the majority of the evaluation datasets, whereas Swin outperformed ViT on the majority of the datasets; however, at a significant cost to latency and larger number of parameters. Thus, it seems that both ResNet50 and ViT may be preferable in some applications that require smaller footprints and faster inference speed. AFR-Net performed the best overall in terms of performance; however, does have a small added latency and increase in number of parameters compared to, for example, ResNet50. However, the significant improvements in performance on many of the datasets seem to justify the added computational costs.

Lastly, the realignment stage utilizing the local embeddings does incur some additional latency, which we will denote as $t_R$. For our implementation, the average value of $t_R$ is 29.36ms. In addition to $t_R$, the realignment stage includes the time required for one additional inference time of the embedding network, $t_I$. However, since we only invoke the realignment stage for a fraction of the total comparisons, $r$, the amortized latency cost, $t_A$, of the realignment is significantly lower and can be computed with the following equation:

\begin{equation}
t_A = r(t_R + 2t_I) + (1-r)t_I
\end{equation}

For example, with a specified range of [0.3, 0.6], the realignment process for AFR-Net is invoked 17.9\% (r=0.179) of the time, on average across all the datasets. Using the inference speed of AFR-Net from Table~\ref{tab:authentication} of 8.42ms, the total cost of AFR-Net$^\dagger$ (AFR-Net with realignment) is 15.18ms.

\subsection{Ablation Analysis}
\label{ablation}
In the abalation study of our AFR-Net model, we analyzed the effects of the loss function (cross entropy vs. ArcFace), training dataset size, use of a spatial transformer network (STN) for spatial alignment, and our realignment strategy using the local feature embeddings. For the ablation on the training dataset size, we compared the performance of our algorithm when trained on only a subset of our full 1.3M training images. Specifically, we created the subset using only the publicly available fingerprint datasets, which included NIST SD 302, IIT Bombay Touchless and Touch-based, ManTech Phase 2, SpoofGAN, and PrintsGAN. This resulted in 760K training images, where 675K of these images are synthetic (from SpoofGAN and PrintsGAN). In comparison, our full training database consists of the same 675K synthetic images plus an additional 540K real fingerprint images.

The results of the ablation study are given in Table~\ref{tab:ablation}. The largest increase in performance is attributed to the use of ArcFace loss rather than a cross entropy loss for supervision. Interestingly, training with ArcFace loss on a subset of only publicly available training data (85K real fingerprints + 675K synthetic compared to our full dataset of 540K real fingerprints + 675K synthetic) achieves competitive recognition performance across all datasets, where really the benefit of additional data is seen in only the cross-sensor and latent matching scenarios. We then saw even further improvements obtained with the incorporation of the spatial alignment network. Finally, we noticed consistent performance improvements across all evaluation datasets with applying our realignment strategy, especially in the more challenging datasets such as NIST SD 302 and FVC 2002 DB3A, which have many partially overlapping fingerprints.

\begin{table*}
\caption{Ablation study for AFR-Net.}
\label{tab:ablation}
\begin{tabular}{c|c|c|c|ccc|ccc|ccccc}
\noalign{\hrule height 1.5pt}
\multirow{3}{*}{\textbf{Loss}} & \multirow{3}{*}{\specialcell{\textbf{\# Images}}} & \multirow{3}{*}{\specialcell{\textbf{STN}}} & \multirow{3}{*}{\specialcell{\textbf{Realign}}} & \multicolumn{6}{c|}{\textbf{TAR (\%) @ FAR=0.1\%}$^*$}                                                                        & \multicolumn{5}{c}{\textbf{TAR (\%) @ FAR=0.01\%}}                                                                                                \\
\cline{5-15}
       &            &                                     &                                                               & \multicolumn{3}{c|}{\textbf{FVC 2002}}                    & \multicolumn{3}{c|}{\textbf{FVC 2004}}                                   & \multirow{2}{*}{\specialcell{\textbf{NIST}\\\textbf{SD14}}} & \multirow{2}{*}{\specialcell{\textbf{NIST}\\\textbf{SD302}}} & \multirow{2}{*}{\specialcell{\textbf{NIST}\\\textbf{SD27}}} & \multirow{2}{*}{\textbf{PolyU}} & \multirow{2}{*}{\textbf{ZJU}} \\
                              &                                  &                               &                                       & \textbf{DB1A}         & \textbf{DB2A}  & \textbf{DB3A}  & \textbf{DB1A}         & \textbf{DB2A}           & \textbf{DB3A}         &                             &                              &                             &                        &                      \\
\noalign{\hrule height 1.0pt}
\specialcell{Cross\\Entropy}                  & 760K                             & No                            & No                                    &      96.93        &    96.50   &    80.21   &     89.61         &       89.14         &      98.50        &           98.23                  &       74.05                       &             21.32                &           69.15             &           87.09           \\
\hline
ArcFace        & 760K                             & No                            & No                                    &       98.46       &    99.86   &     92.50  &      99.00        &     98.61           &       99.54       &          99.63                   &             92.04                 &          39.53                   &           91.3             &            97.61          \\
\hline
ArcFace        & 1.3M                             & No                            & No                                    & 99.79        & 99.82 & 97.82 & 99.82        & 98.68          & 99.93        & 99.89                       & 94.82                        & 58.14                       & 97.00                     & 98.66                \\
\hline
ArcFace        & 1.3M                             & Yes                           & No                                    & 99.86        & \textbf{99.96} & 98.43 & \textbf{100} & 99.36          & \textbf{100} & \textbf{99.93}              & 95.46                        & \textbf{63.18}              & 98.23                  & 98.68                \\
\hline
ArcFace        & 1.3M                             & Yes                           & Yes                                   & \textbf{100} & \textbf{99.96} & \textbf{99.36} & \textbf{100} & \textbf{99.64} & \textbf{100} & \textbf{99.93}              & \textbf{96.11}               & \textbf{63.18}              & \textbf{98.73}         & \textbf{98.7}     \\
\noalign{\hrule height 1.5pt}
\multicolumn{15}{p{0.95\linewidth}}{$^*$ Following the FVC protocol of 2,800 genuine pairs and 4,950 imposter pairs.}\\
\end{tabular}
\end{table*}

\section{Discussion}
In this section, we discuss some remaining failure cases of our model and some possible future extensions to mitigate those. We also investigate the robustness of our model to partial fingerprints by manually generating affine and occlusion deformations (of varying magnitudes) across multiple fingerprint datasets.

\subsection{Failure Case Analysis}
\begin{figure}
\centering
\includegraphics[width=\linewidth]{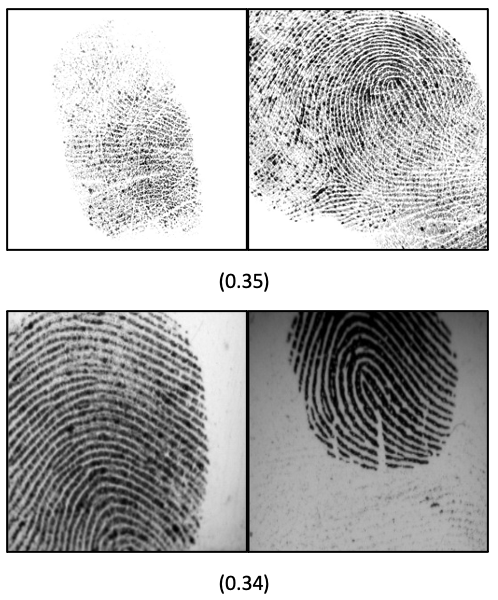} 
\caption{Example fingerprint pairs that failed to match successfully by the AFR-Net model. The top pair is from NIST SD 302 and the bottom pair is from FVC 2002 DB2A. Similarity scores are given below each pair, both below the match threshold of 0.36.}
\label{fig:failures}
\end{figure}

Two example fingerprint image pairs that failed to be successfully matched by our AFR-Net model are shown in Figure~\ref{fig:failures}. As demonstrated with these representative examples, the majority of the failure cases can be attributed to one of two factors: i.) extremely poor image quality or ii.) very little overlap between the images. The first cause can be avoided by implementing a quality check into the algorithm, whereas the second cause may be more difficult to detect and/or avoid in a practical system (especially one operating with a limited acquisition aperture). Our realignment strategy is effective at improving partial overlap pairs; however, when the amount of overlap is severe, such as example (b) in Figure~\ref{fig:failures}, the model may still fail.

\subsection{Robustness to Occlusions and Affine Transformations}
\begin{figure*}[!t]
\centering
\includegraphics[width=\linewidth]{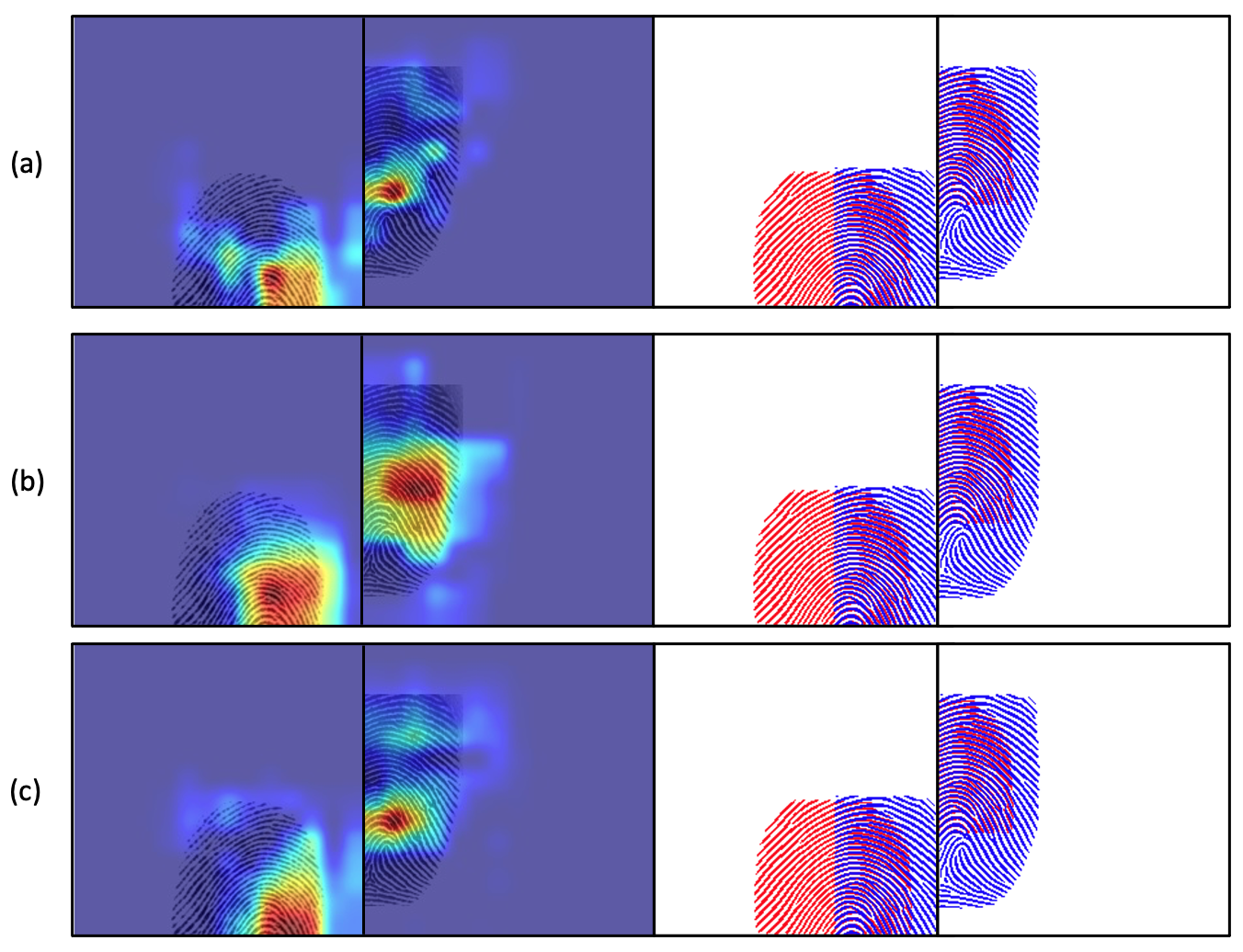} 
\caption{Saliency maps for a partial fingerprint pair from FVC 2002 DB1A computed with (a) ResNet50, (b) ViT, and (c) AFR-Net models. The reddish regions represent regions of the fingerprint where the similarity score drops the most when occluded (indicating it's importance), whereas the blue regions represent the regions with high similarity scores even with that region occluded (indicating low importance). On the right, the ridge structures of each fingerprint are overlaid to highlight the overlapping area. (Best viewed in color).}
\label{fig:saliency}
\end{figure*}

To help understand the difference between CNN-based and attention-based embeddings, we conducted an experiment to visualize the saliency maps of each model on pairs of partial fingerprints. Specifically, we scan a patch of 16x16 background pixels (with a stride of 16) across one image in the pair and compute the resulting similarity scores, which we use to draw a heatmap of salient regions for each patch in the image. We repeat this process for the other image and overlay the heatmaps onto the original images. Due to space limitations, we just show one representative example in Figure~\ref{fig:saliency}, along with the ridge overlays of the two images to better visualize the overlapping regions. Comparing the saliency maps of ResNet50 ((a) in Figure~\ref{fig:saliency}) to ViT ((b) in Figure~\ref{fig:saliency}), it seems that occluding some areas of the fingerprint has more of an effect on varying the similarity scores for ViT than it does for ResNet50. This suggests that ViT is placing more weight on specific regions of the fingerprint, compared to ResNet50 which may be using more of the fingerprint area for its prediction. Finally, comparing both saliency maps with the saliency map of AFR-Net (shown in (c) of Figure~\ref{fig:saliency}), we can see that AFR-Net exhibits characteristics of both models.

\begin{figure}
\centering
\includegraphics[width=\linewidth]{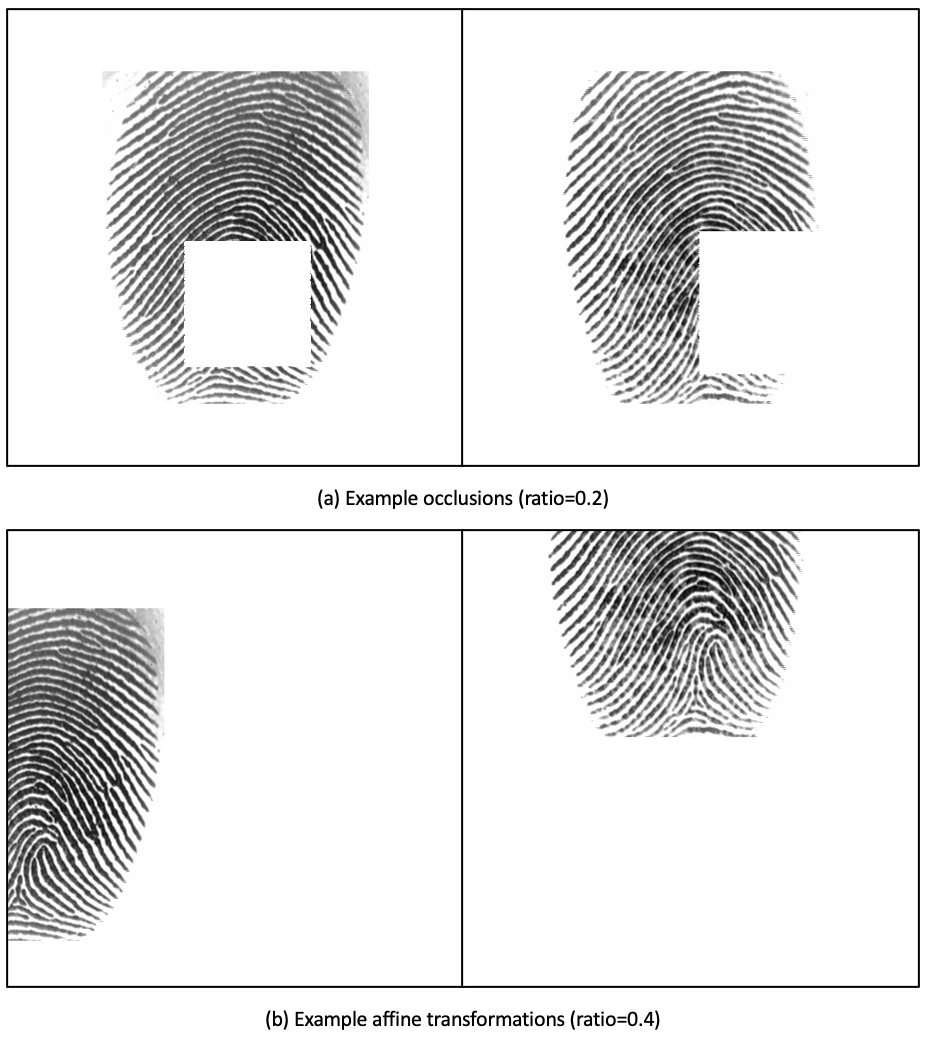} 
\caption{Example manual occlusions and affine transformations to generate challenging, partial fingerprint pairs.}
\label{fig:occlusion}
\end{figure}

We performed two additional experiments using manual occlusions and affine transformations to generate increasingly occluded and cut-off partial fingerprints and plotted the performance of each model vs. the amount of degradation. Specifically, we generated random occlusions and affine transformations at five different ratios, corresponding to the percentage of the fingerprint area being obscured. We repeated the experiment 5 times at each ratio and recorded the average performance of each model. For reference, example images with random affine transformations and occlusions at ratios of 40\% and 20\%, respectively, are shown in Figure~\ref{fig:occlusion}.

\begin{figure}
\centering
\includegraphics[width=\linewidth]{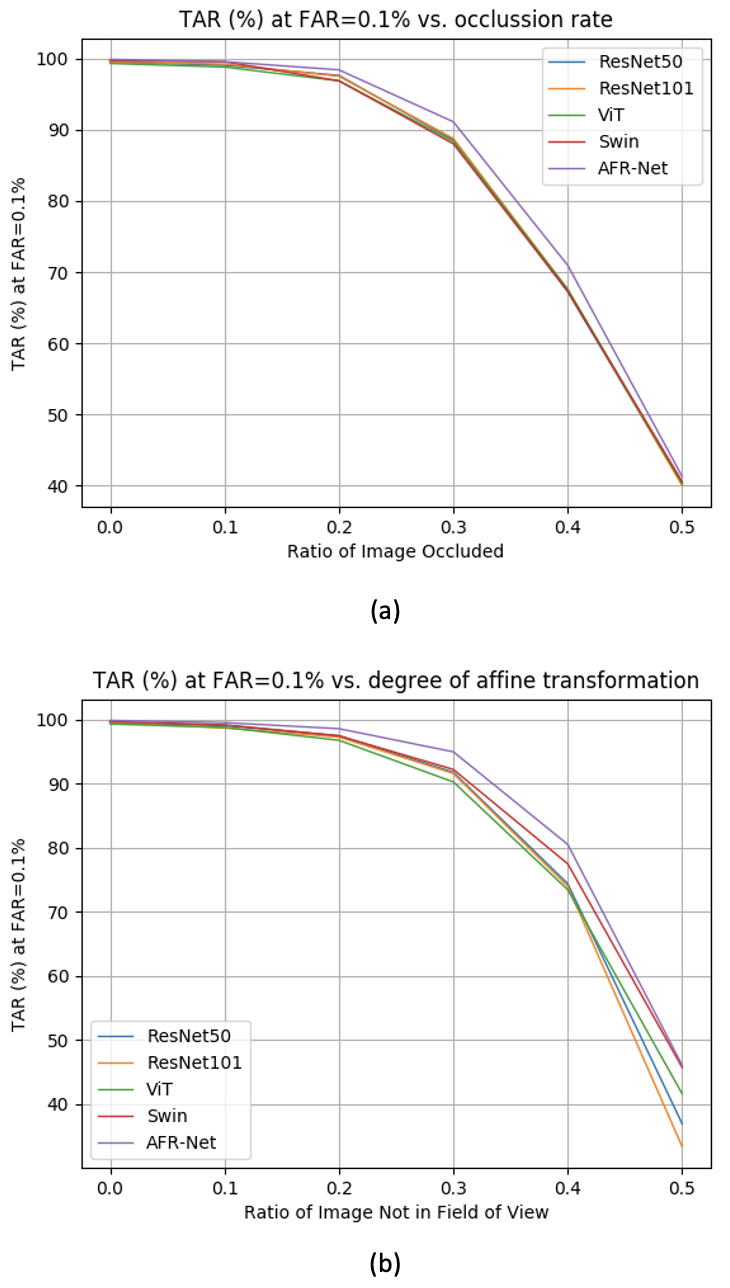} 
\caption{Comparison in TAR (\%) at FAR=0.1\% of each model under increasing degrees of random (a) occlusion and (b) affine transformations.}
\label{fig:tar_vs_occlusion}
\end{figure}

According to Figure~\ref{fig:tar_vs_occlusion}, the ResNet models appear to have a slight edge over the attention-based models (ViT and Swin) when subjected to random occlusions, whereas the ViT and Swin models show more robustness to severe degrees of affine transformations compared to the ResNet models. However, AFR-Net shows the best robustness to both occlusions and affine transformations, underscoring the benefit of merging the two complimentary networks.

\section{Conclusion}
In this paper, we evaluated attention-based fingerprint recognition networks against competitive CNN baselines and a state-of-the-art commercial fingerprint recognition system, Verifinger v12.3, and showed that our combined architecture, AFR-Net (Attention-Driven Fingerprint Recognition Network), outperforms all of the baselines in the majority of the evaluation datasets. These evaluations included intra-sensor, cross-sensor, contact to contactless, and latent fingerprint matching scenarios. Furthermore, we introduced a realignment stage using the correspondence between local embeddings extracted from intermediate feature maps of two fingerprint images which consistently improved the performance across all the models, especially in challenging cases (e.g., partial overlap between the fingerprint images). This realignment strategy requires no additional training and can be applied as a wrapper to any deep learning network (CNN or attention-based). It also serves as an explainable visualization of the corresponding regions of two fingerprint images as ascertained by the network. Future work will aim at improving the realignment strategy to reduce the latency introduced by the current brute force correspondence implementation. We will investigate the use of attention and/or graphical neural networks for this purpose in order to more intelligently aggregate two sets of local embeddings.


\ifCLASSOPTIONcaptionsoff
  \newpage
\fi



\bibliographystyle{IEEEtran}
\bibliography{cite}
%



%

\begin{IEEEbiography}[{\includegraphics[width=1in,height=1.25in,clip,keepaspectratio]{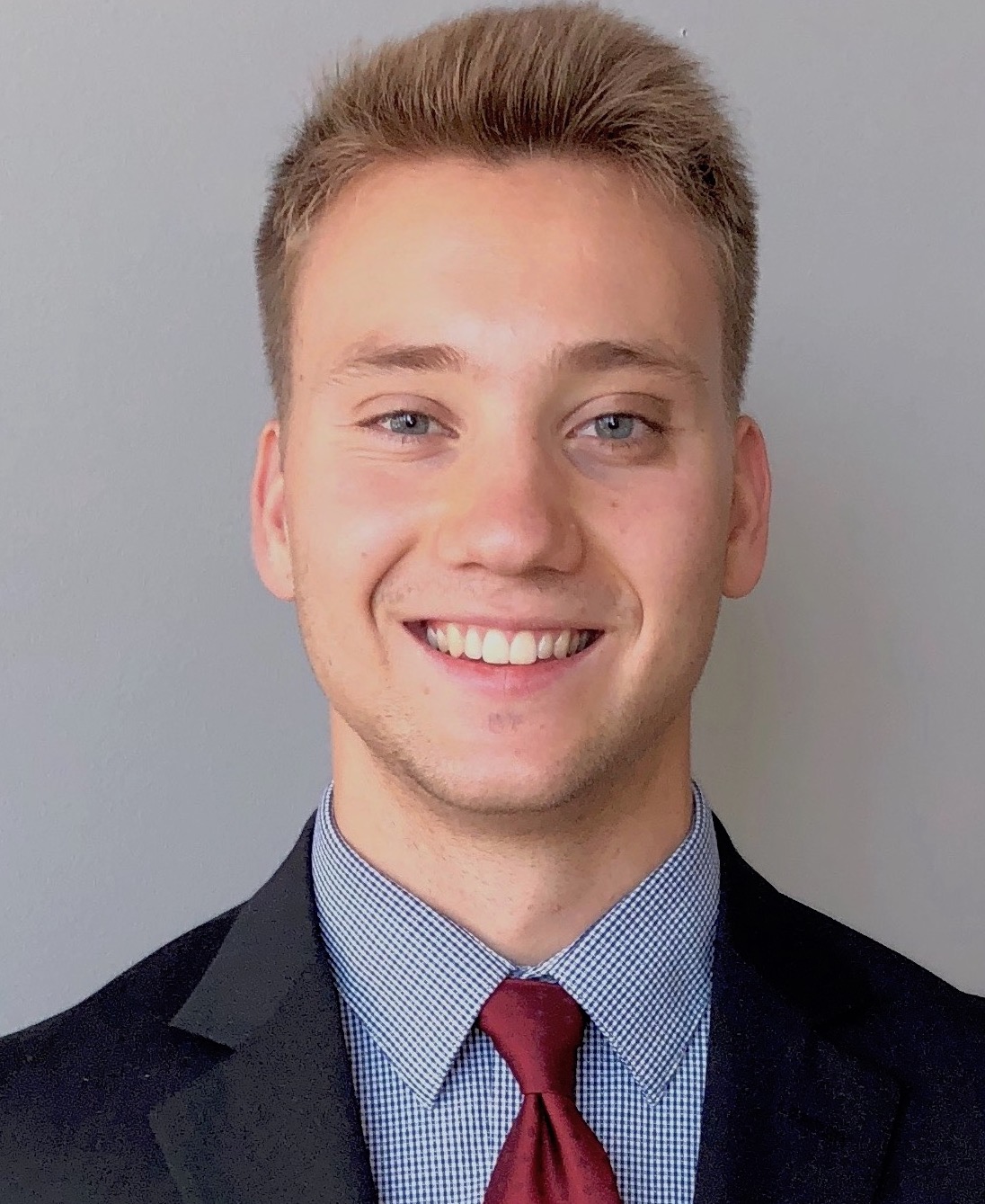}}]{Steven A. Grosz}
received his B.S. degree with highest honors in Electrical Engineering from Michigan State University, East Lansing, Michigan, in 2019. He is currently a doctoral student in the Department of Computer Science and Engineering at Michigan State University. His primary research interests are in the areas of machine learning and computer vision with applications in biometrics.
\end{IEEEbiography}

\begin{IEEEbiography}[{\includegraphics[width=1in,height=1.25in,clip,keepaspectratio]{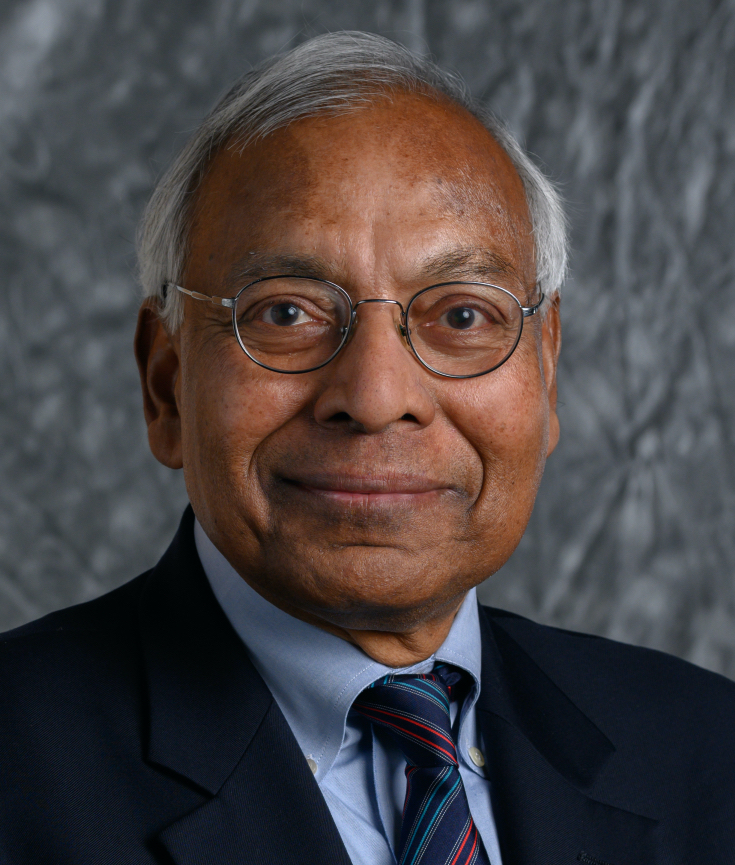}}]{Anil K. Jain}
Anil K. Jain is a University distinguished professor in the Department of Computer Science and Engineering at Michigan State University. His research interests include pattern recognition and biometric authentication. He served as the editor-in-chief of the IEEE Transactions on Pattern Analysis and Machine Intelligence and was a member of the United States Defense Science Board. He has received Fulbright, Guggenheim, Alexander von Humboldt, and IAPR King Sun Fu awards. He was elected to the National Academy of Engineering, the Indian National Academy of Engineering, the World Academy of Sciences, and the Chinese Academy of Sciences.
\end{IEEEbiography}






\clearpage
\appendix

\begin{figure}[h!]
\centering
\includegraphics[width=\linewidth]{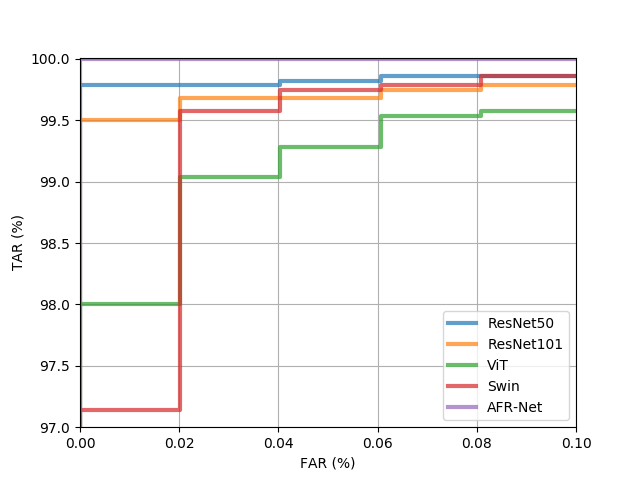} 
\caption{ROC curve for FVC 2002 DB1A.}
\label{roc:fvc2002db1a}
\end{figure}

\begin{figure}[h!]
\centering
\includegraphics[width=\linewidth]{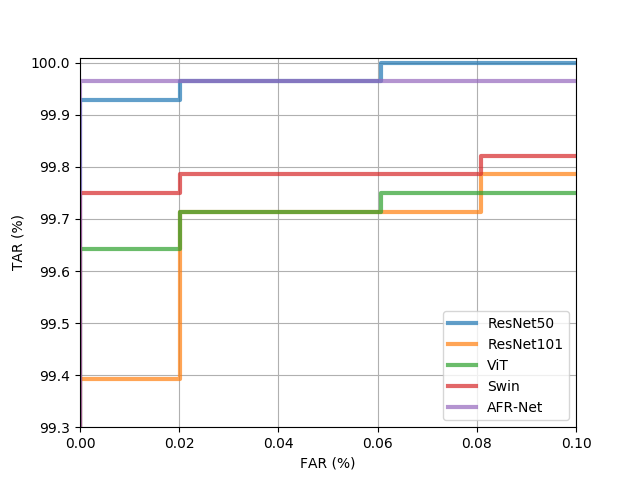} 
\caption{ROC curve for FVC 2002 DB2A.}
\label{roc:fvc2002db2a}
\end{figure}

\begin{figure}[h!]
\centering
\includegraphics[width=\linewidth]{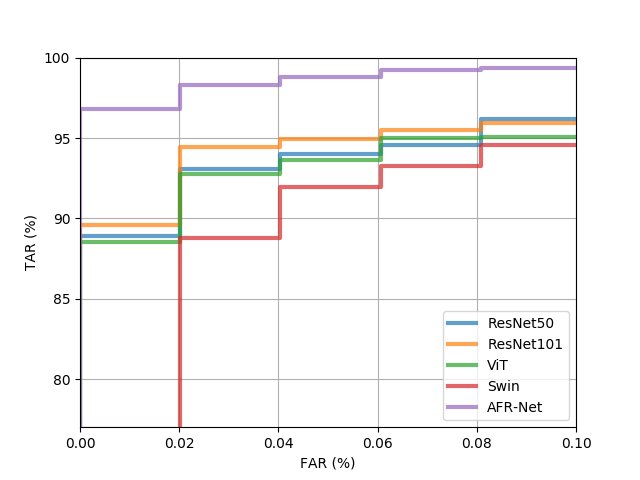} 
\caption{ROC curve for FVC 2002 DB3A.}
\label{roc:fvc2002db3a}
\end{figure}

\begin{figure}[h!]
\centering
\includegraphics[width=\linewidth]{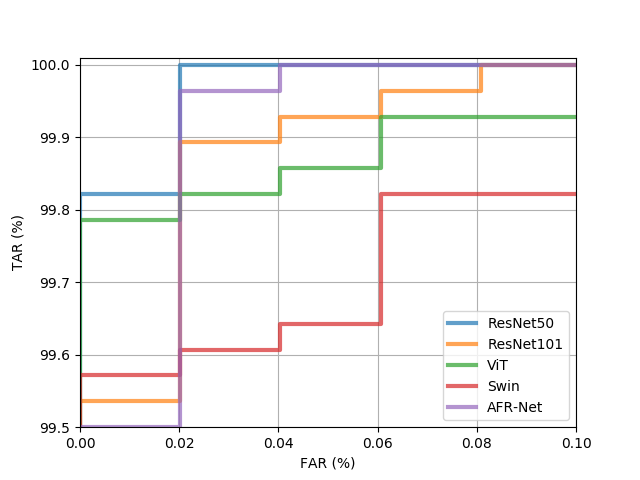} 
\caption{ROC curve for FVC 2004 DB1A.}
\label{roc:fvc2004db1a}
\end{figure}

\begin{figure}[h!]
\centering
\includegraphics[width=\linewidth]{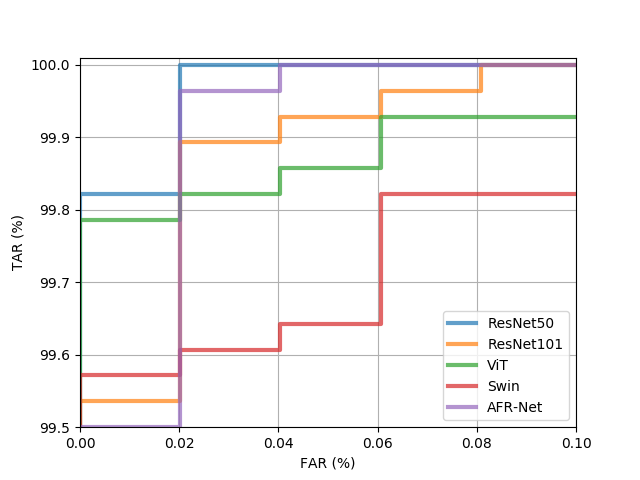} 
\caption{ROC curve for FVC 2004 DB1A.}
\label{roc:fvc2004db2a}
\end{figure}

\begin{figure}[h!]
\centering
\includegraphics[width=\linewidth]{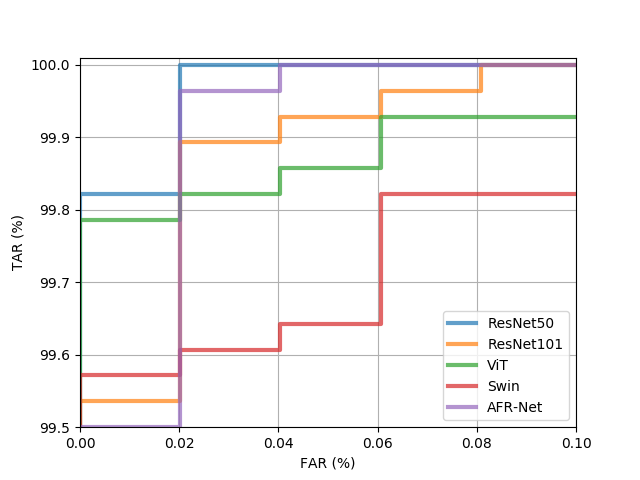} 
\caption{ROC curve for FVC 2004 DB1A.}
\label{roc:fvc2004db3a}
\end{figure}

\begin{figure}[h!]
\centering
\includegraphics[width=\linewidth]{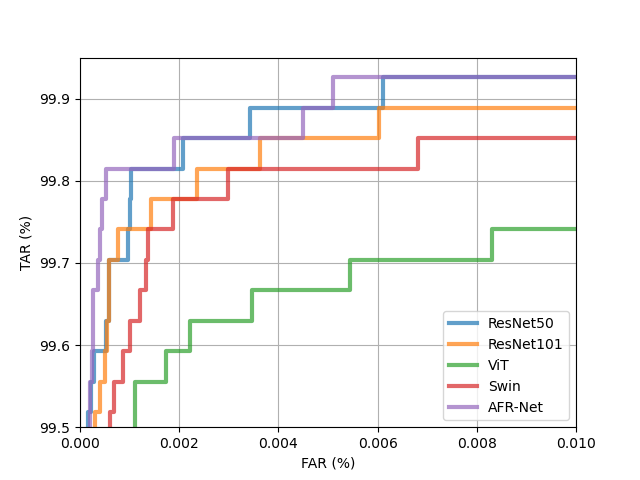} 
\caption{ROC curve for NIST SD 14.}
\label{roc:nistsd14}
\end{figure}


\begin{figure}[h!]
\centering
\includegraphics[width=\linewidth]{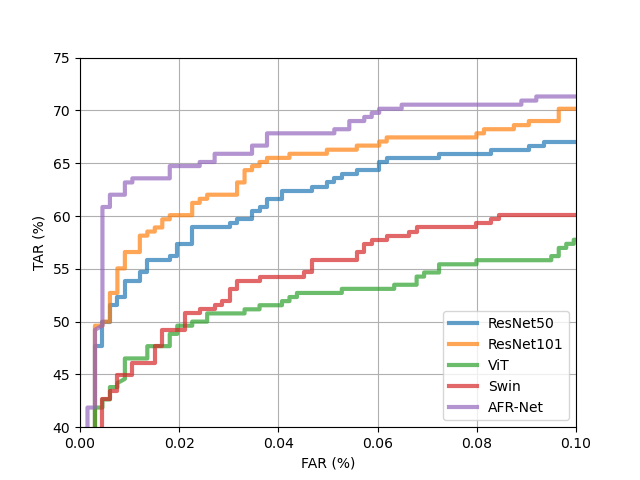} 
\caption{ROC curve for NIST SD 27.}
\label{roc:nistsd27}
\end{figure}

\begin{figure}[h!]
\centering
\includegraphics[width=\linewidth]{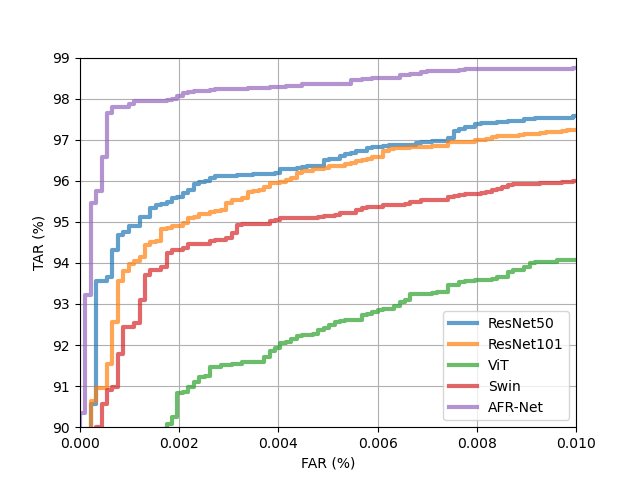} 
\caption{ROC curve for PolyU Contactless 2D to Contact-based 2D Database.}
\label{roc:polyu}
\end{figure}

\begin{figure}[h!]
\centering
\includegraphics[width=\linewidth]{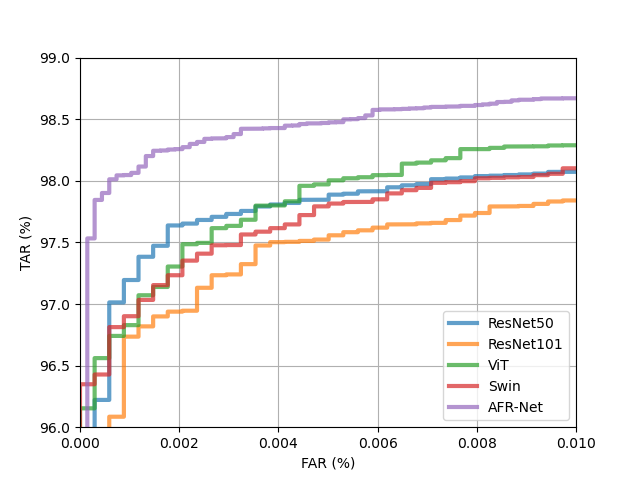} 
\caption{ROC curve for ZJU Finger Photo and Touch-based Database.}
\label{roc:zju}
\end{figure}

\end{document}